\documentclass[a4paper,conference]{IEEEtran}
\ifCLASSINFOpdf
\else
\fi
\usepackage{amsmath}
\newcommand\norm[1]{\lVert#1\rVert}

\usepackage{graphicx}
\usepackage{kotex}
\usepackage{indentfirst}
\usepackage{amssymb}
\usepackage{pifont}
\usepackage[noadjust]{cite}
\usepackage{xcolor}

\usepackage{array}
\usepackage{multirow,makecell}
\usepackage{algorithm} 
\usepackage{algpseudocode} 
\begin{document}
%
\title{Multi-Contextual Predictions with \\ Vision Transformer for Video Anomaly Detection}

\author{\IEEEauthorblockN{Joo-Yeon Lee$^1$, Woo-Jeoung Nam$^2$, Seong-Whan Lee$^{1,2}$}
\IEEEauthorblockA{1. Department of Artificial Intelligence, Korea University\\
2. Department of Computer and Radio Communications Engineering, Korea University \\
Email: \{jooyeon\_lee, nwj0612, sw.lee\}@korea.ac.kr}\\
}


\maketitle

\begin{abstract}

Video Anomaly Detection(VAD) has been traditionally tackled in two main methodologies: the reconstruction-based approach and the prediction-based one. As the reconstruction-based methods learn to generalize the input image, the model merely learns an identity function and strongly causes the problem called \textit{generalizing issue}. On the other hand, since the prediction-based ones learn to predict a future frame given several previous frames, they are less sensitive to the \textit{generalizing issue}. However, it is still uncertain if the model can learn the spatio-temporal context of a video. Our intuition is that the understanding of the spatio-temporal context of a video plays a vital role in VAD as it provides precise information on how the appearance of an event in a video clip changes. Hence, to fully exploit the context information for anomaly detection in video circumstances, we designed the transformer model with three different contextual prediction streams: masked, whole and partial. By learning to predict the missing frames of consecutive normal frames, our model can effectively learn various normality patterns in the video, which leads to a high reconstruction error at the abnormal cases that are unsuitable to the learned context. To verify the effectiveness of our approach, we assess our model on the public benchmark datasets: USCD Pedestrian 2, CUHK Avenue and ShanghaiTech and evaluate the performance with the anomaly score metric of reconstruction error. The results demonstrate that our proposed approach achieves a competitive performance compared to the existing video anomaly detection methods.
\end{abstract}

%
\IEEEpeerreviewmaketitle

\section{Introduction}
Video anomaly detection (VAD), which detects abnormal events that do not conform to a defined normal pattern of video content (e.g., a running person in a surveillance video), has attracted significant attention in many real-world applications.
VAD is a very challenging task because abnormal events occur much less frequently than normal events in the real world, and the definition of abnormal behavior depends on circumstances and context.
Therefore, VAD methods aim to learn normal patterns of frames of training video in an unsupervised manner and then identify the frame of testing video as abnormal if it varies from the established normal patterns beyond some allowable threshold.

 \begin{figure}[!t]
 \centering
\includegraphics[width=3.2in]{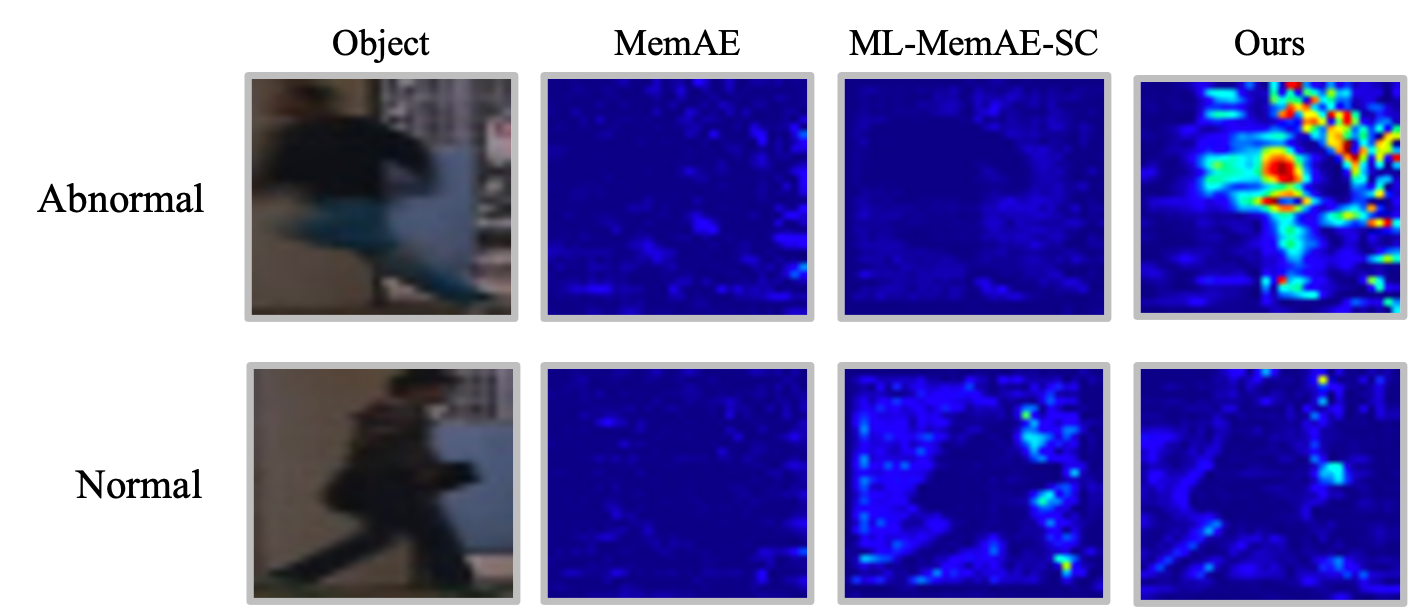}
\caption{Visualization of the reconstruction error maps which are obtained with the \begin{math}\ell_{2}\end{math} distance between the ground-truth frames and the reconstructed frames using object-level appearance frames. Autoencoders with memory modules (MemAE\cite{gong2019memorizing}, ML-MemAE-SC in \cite{liu2021hybrid}) are visualized in the second and third column. The fourth column shows the error map of our model. The first and second row represent abnormal(running) and normal(walking) frames. The lighter the color, the greater the error value.}
\label{fig_sim}
\end{figure}

Recently, deep learning methods have shown great success in VAD\cite{gong2019memorizing,liu2021hybrid,feng2021convolutional,hasan2016learning,liu2018future,park2020learning,yu2020cloze,cai2021appearance,hinami2017joint,morais2019learning,szymanowicz2021x, ouyang2021video}. There have been two main branches in VAD methodologies: the reconstruction-based approach and the prediction-based approach. In the reconstruction-based methods, deep-learning models such as generative networks\cite{hasan2016learning,goodfellow2014generative}, learn to recognize normal patterns by reconstructing frames of normal training videos and use the reconstruction errors between the input frame and reconstructed frames as an anomaly score. 
If the reconstruction error is larger than a given threshold, the frames are considered to be abnormal, since the model does not learn abnormal frames at train time thus the model cannot reconstruct abnormal frames effectively.
In the future frame prediction-based methods\cite{liu2018future,cai2021appearance,yu2020cloze,feng2021convolutional,yuan2021transanomaly}, on the other hand, the generative models learn to predict the next frame based on a given set of previous frames. By predicting previously unseen frames, they become able to learn enhanced feature representations, which typically leads to the better anomaly detection results than the reconstruction-based ones.
Both of these two methods assume that the abnormal frames are not soundly reconstructed as the training process only involves with normal frames. However, the deep generative models tend to merely learn an identity function, not selectively learning normality patterns of the videos. Hence even when the abnormal frames are fed into the model in current VAD methodologies, the model simply regenerates the frame, not making it closer to the learned normality patterns. Due to this high reconstructing performance, a problem called \textit{generalizing issue} occurs.

To address the aforementioned problem, Autoencoders (AE) with a memory module\cite{gong2019memorizing,park2020learning,cai2021appearance} have been extensively studied. These approaches focus on increasing the difference between an abnormal input frame and a reconstructed frame by replacing the test frame with memory slots that store prototypical normal patterns in training videos. Nevertheless, these methods still could not effectively resolve the issue, as there is no guarantee that the various normality patterns are stored in the memory module.
In Fig. 1, we show the visualized reconstruction error maps on object-level appearance images of VAD dataset in the memory module based methods(MemAE\cite{gong2019memorizing} and ML-MemAE-SC in HF2-VAD\cite{liu2021hybrid}). The difference between the reconstruction error of the normal data and the abnormal data counterpart is not noticeably large. Since a running person(an abnormal behavior) is properly reconstructed in these memory module methods, the reconstruction error is restricted to be small, causing the inability to distinguish between normal and abnormal events. 


Based on the above observation, we propose a VAD method that is free from the dependency on the memory modules. Instead, we apply self-supervisory simultaneous prediction to enhance the understanding of how the appearance of an event in the video clip has changed. To be concrete, the model predicts the masked frames among the consecutive frames at once, such that the model can fully consider the context information of the video unlike the existing single-frame prediction approaches.
This simultaneous self-supervisory future frame prediction helps the model to generate the prediction outputs which are closer to the training data distribution, not simply learning the identity function and consequentially alleviating the \textit{generalizing issue}. 

There is another additional branch of the model that processes the optical flow of the video along with the contextual-appearance. 
Since abnormal events are usually viewed different from the normality in appearance and motion, we detect the anomalies in terms of the flow and the appearance separately and integrate results in the end. We construct the motion module as a convolutional autoencoder(CAE)model that uses the optical flow data as an input. 
In summary, our contributions are as follows.
\begin{itemize}
    \item We apply the ViT model to the VAD task with the pixel-level prediction, which leads to the learning of the features with contextual information. 
    We carefully revisit the reconstruction error map to confirm that model allows to learn normality as our assumption.
   
    \item We design multi-contextual predictions that predict the missing components of the consecutive normal frames in three ways: whole, masked and partial situations, simultaneously. Our approach to predict the multi-contextual frames in different streams leads to a clear distinction between normal and abnormal events for object-level VAD tasks, attaining better performance than the memory module autoencoder.
    \item Our framework combined with flow reconstruction shows the synergy to increase the performance, resulting in high reconstruction errors in the abnormal events. We evaluate our method on the public datasets: USCD Pedestrian 2, CUHK Avenue and ShanghaiTech with demonstrating competitive performance compared to the existing methods.
\end{itemize}

\begin{figure*}[!t]
\includegraphics[width=\textwidth,height=190pt]{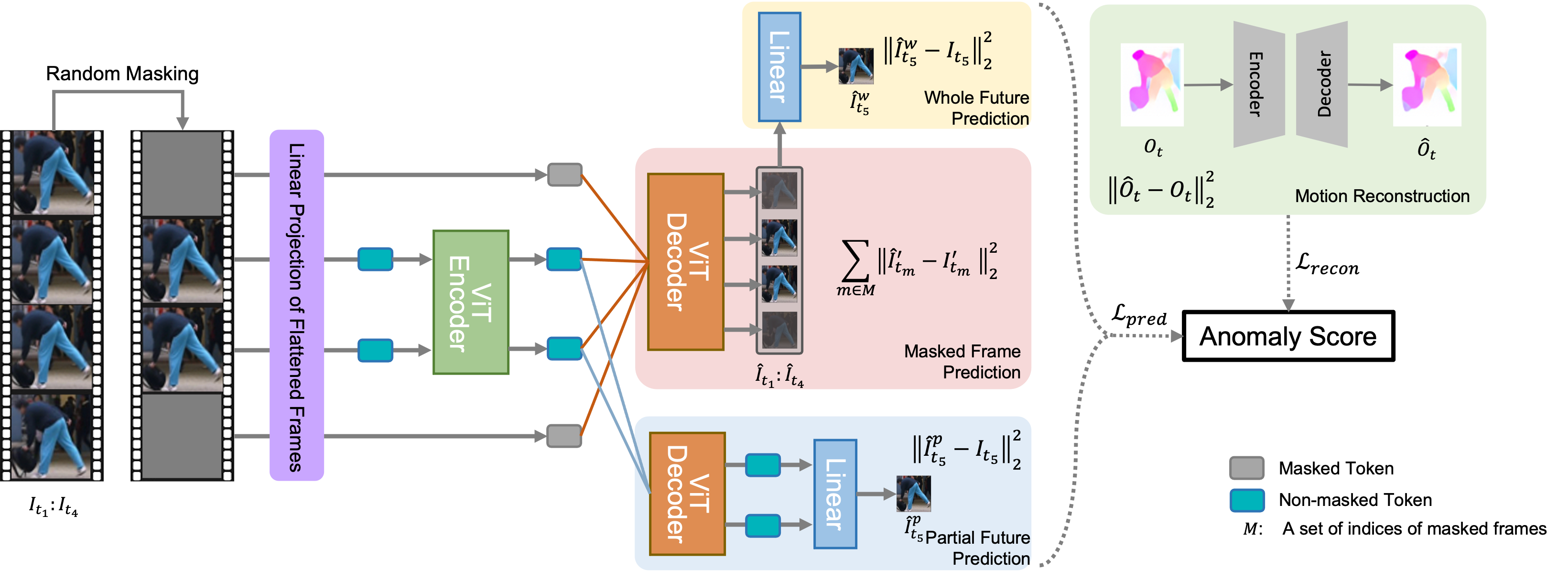}
\caption{
Overview of the proposed framework. We randomly set mask on the consecutive object-level input frames $I_{t_{1}}:I_{t_{4}}$. Each frame is tokenized as a vector form through the patch embedding process. Only the tokens of non-masked frames are passed through the ViT encoder, and then fed into the two separate ViT decoders based on the presence of the mask on the tokens. Our framework computes three different contextual predictions from the input frames: i) whole, ii) partial and iii) masked frame prediction. 
The first ViT decoder of the masked frame prediction module(the pink background) utilizes both of the masked tokens and non-masked ones to predict the masked frames, which are depicted translucently in the figure. The output of the first ViT decoder is used for predicting the next frame through the linear embedding in the whole future prediction module(the yellow background). The second ViT decoder of the partial future prediction module(the blue background) learns to predict the next frame using the partial inputs that only contain non-masked tokens. The motion reconstruction module(the green background) learns the normality of the motions while reconstructing the optical flow of the $t$-th frame.
The losses from the three different prediction modules are integrated with the loss from the motion reconstruction module to finally compute the anomaly score.}
\label{fig_sim}
\end{figure*}

\section{Related work}
\subsection{Video Anomaly Detection}
Beyond the Machine learning\cite{lee2003pattern,xi2002facial,lim2000text,lee1990translation} methods, the deep-learning methods\cite{vaswani2017attention,carion2020end,yang2007reconstruction,lee1999integrated} are used in many fields.
Among them, many deep-learning-based VAD methods aim to learn normality patterns by training only normal samples with a deep-learning model\cite{gong2019memorizing,liu2021hybrid,feng2021convolutional,hasan2016learning,liu2018future,park2020learning,yu2020cloze,cai2021appearance,hinami2017joint,morais2019learning,szymanowicz2021x}.
Reconstruction and prediction-based methods are currently the most widely used methods. Reconstruction-based methods\cite{gong2019memorizing,hasan2016learning}, learn normality patterns by reconstructing training inputs using a generative model such as an autoencoder (AE).

The prediction-based method\cite{liu2018future}, is also similar to reconstruction-based methods, but predicts the unseen next frame.
These two approaches assume that at test time, abnormal frames that are not learned can not be properly predicted because the training samples consist only of normal frames.
Hence, the reconstruction error of abnormal frames has to be larger than that of normal frames.
However, this assumption is not always correct because of the high capacity of generalizing ability in generalization models, which can reconstruct even non-learned abnormal frames well. This problem called \textit{generalizing issue}.

To alleviate this \textit{generalizing issue}, an AE with a memory module\cite{gong2019memorizing,park2020learning,cai2021appearance}, has been proposed in recent years. These methods are described in \uppercase\expandafter{\romannumeral1}.
   
\subsection{Vision Transformer}
ViT\cite{dosovitskiy2020image} follows the structure of the original Transformer\cite{vaswani2017attention} used in the Natural Language Processing and has the advantage of being able to learn sequential information.
ViT splits an image into multiple patches and linearly embeds each of them with position information. The outputs are then fed into the Transformer encoder which contains a self-attention module. The module obtains the relationship between the current patch and all given patches, and thus encodes the current patch considering information on all given patches. After the feed-forward network (FFN) and the residual connection, these encoded patches are then used to perform the classification task. 


ViT typically shows a great dependency on the data abundance as it suffers from the inductive biases caused by the inability to learn various localities, resulting in ignoring the overlapping parts between the patches. Hence, one may assume applying ViT for VAD task is not appropriate, as the amount of VAD datasets is not that large compared to the other tasks such as image classification or object detection. Several approaches exploiting Transformer to the anomaly detection\cite{yuan2021transanomaly,feng2021convolutional} are also only based on the Transformer with convolutional layers, not ViT. However, our approach proves that ViT can be successfully trained to detect anomalies in video even without a huge amount of data. 
Our proposed method is less sensitive to the problem of inductive bias since a single patch means one image in our model, so that there is no need to consider the overlap between the patches.

Moreover, we are inspired by MAE\cite{he2021masked} for the simultaneous multi-frame prediction.
MAE\cite{he2021masked} generalizes an image by leveraging ViT. It firstly splits an image into multiple patches, and randomly mask some of the patches. The masked patches are then reconstructed to learn an enhanced feature representation. Although it reconstructs images at the pixel-level, which does not comprise semantic entities, it can effectively generalize images with semantics owing to the rich hidden representations obtained.

\section{Method}
The proposed framework is mainly composed of the contextual-appearance module and the motion reconstruction module as shown in Fig.2.
The contextual-appearance module is trained only on the normal training samples composed of the object-level frames extracted from the training videos. Likewise, the motion reconstruction module is trained only on the optical flow of the two consecutive normal samples of the object-level frames. During the inference, the object-level frames \begin{math}t_{1}:t_{4}\end{math} are randomly masked, according to the preset masking ratio. 
Our framework computes the three different contextual predictions from the input frames: i) masked, ii) whole and iii) partial frame prediction. The first ViT decoder of the masked frame prediction module utilizes both of the masked tokens and non-masked ones to predict the masked frames. The output of the first ViT decoder is used for predicting the next frame through the linear embedding in the whole future prediction module. The second ViT decoder of the partial future prediction module learns to predict the next frame using the partial inputs that only contain non-masked tokens. The motion reconstruction module learns the normality of the motions while reconstructing the optical flow of the $t$-th frame. The losses from the three different prediction modules are integrated with the loss from the motion reconstruction module to finally compute the anomaly score. If this error is larger than a given threshold, the test sample is defined as abnormal.
In the following section, we introduce each module and the method of the framework more specifically.

\subsection{Contextual-appearance module with Transformer}
A video is a bundle of changes in between the frames by definition, hence the change of the appearance of an event is a primary key for comprehending the video. To reflect this intuition to our framework, we introduce a simultaneous multi-frame prediction to leap the model's understanding of the spatio-temporal context. To avoid depending on a number of the deep neural networks to process the multiple frames at once as in the previous work\cite{yu2020cloze}, we utilize ViT model.

The future frame prediction is performed in the two different manners: whole and partial, whereas the masked frame prediction reconstructs the current masked frames. In the whole future prediction, the model predicts the future frame by leveraging the relationships between the non-masked tokens and the masked tokens. 
As the non-masked tokens are composed of the parameters that are only trained on the normal samples, the predicted frame can effectively reflect the normality. 
In the partial future prediction, while the model is restricted to perform a more challenging task of predicting the next frame based only on the given incomplete information, the latent feature space of the model gets improved.

We define the set of indices of all tokens and the masked tokens as \begin{math}S,M\end{math} where $M\in S$.
\begin{math}E,D^{w},D^{p}\end{math} denotes the functions of a deep ViT encoder and the two shallow ViT decoder for whole/partial prediction, respectively. LN means a linear layer, and $\phi^{w}, \phi^{p}$ indicate the linear functions for the whole and partial future prediction.
Inputs with the size of $3\times 32\times 32$ are flattened and linearly embedded, then the tokens  \begin{math}q_{i}\in \mathbb{R}^{C}\end{math} are the output, where \begin{math}C\end{math} denotes the dimension of the token.
Only the non-masked tokens are encoded by the encoder:
\begin{equation}LN(E(q_{i\in M^{c}}))\mapsto R, r_{i} \in R\cup {\{q_{j} | j\in M\}} \end{equation}
Then, non-masked tokens and the asked tokens that consist of the learned variables to generalize the masked frames are fed into the first branch decoder while considering the information of all masked and non-masked tokens. Here, only the masked frames are reconstructed in the the decoder and we define this as the masked frame prediction.
\begin{equation}LN(D^{w}(r_{i\in S}))\mapsto \hat{I'}_{t_{i\in M}}\end{equation}
As all the masked and the non-masked predictions are linearly embedded and used to predict the next frame, Eq.3 denotes the whole future prediction.
\begin{equation}\phi^{w}(\hat{I'}_{t_{i\in S}})\to \hat{I^{w}_{t_{5}}}\end{equation}
In the second branch, only the non-masked tokens are fed into the decoder to predict the next frame, and we define  Eq. 4 as the partial future prediction because only the information of the non-masked tokens is used for the prediction.
\begin{equation}\phi^{p}(D^{p}(r_{i\in M^{c}}))\to \hat{I^{p}_{t_{5}}}\end{equation}

Our objective is to minimize the pixel-wise \begin{math}\ell_{2}\end{math} distance between the predicted frames and their respective ground truths:
\begin{equation}
\begin{split}
\mathcal{L}_{pred} = 
\norm{\hat{I}^{w}_{t_{5}}-I_{t_{5}}}^{2}_{2}
+\norm{\hat{I}^{p}_{t_{5}}-I_{t_{5}}}^{2}_{2}
+\sum_{m\in M}\norm{\hat{I'}_{t_{m}}-I'_{t_{m}}}^{2}_{2} 
\end{split}
\end{equation}
We define \begin{math} \hat{I}^{w}_{t_{5}} \end{math} as the whole future prediction, \begin{math} \hat{I}^{p}_{t_{5}}\end{math} as the partial future prediction. 
If \begin{math}m\in M\end{math}, \begin{math} \hat{I'}_{t_{m}}\end{math} means the masked frame prediction. Also, \begin{math} I_{t_{5}} \end{math} denotes the ground truth of the next frame, and \begin{math} I'_{t_{m}} \end{math} indicates the ground truth of randomly masked frames.


\subsection{Motion module}
It is important to extract motion information from video contents\cite{sun2018optical,choutas2018potion,lee2015motion,lee2022human,lee2021uncertainty}.
Our intuition is that optical flow can be helpful for detecting the abnormal events as the optical flow expresses the speed and the direction of an object of the training samples. By introducing the motion information to our model, the framework can effectively detect the objects moving in the unseen speed and the directions in the normal training samples.
We used a CAE to reconstruct the optical flow at the object-level.
Since the optical flow contains the motion information of an object, the normality of the motions can be learned by simply performing a reconstruction task.
Similar to the contextual-appearance module, we used the \begin{math}\ell_{2}\end{math} distance as a measure of abnormality:
 \begin{equation}\mathcal{L}_{recon} = \norm{\hat{O}_{t}-O_{t}}^{2}_{2} \end{equation}
We defined \begin{math}\hat{O}_{t}\end{math} as the reconstructed flow and \begin{math}O_{t}\end{math} as the ground-truth flow. Here, a flow is calculated as the difference between the two consecutive video frames.

\subsection{Anomaly score}
We utilize the complementary points between the optical flow and the appearance frame to identify abnormal events. Loss values from both of the contextual-appearance modules and the motion reconstruction module prevent the modules being biased to the low reconstruction error in the abnormal situations.
The loss function for the anomaly score is as follows:
 \begin{equation}\mathcal{S} =\lambda_{a}\mathcal{L}_{pred}+\lambda_{o}\mathcal{L}_{recon}. \end{equation}
Here, the final anomaly score $\mathcal{S}$ is the weighted sum of the reconstruction errors of the optical flow and the reconstruction error of appearance.
The details of the hyperparameters are described in the section \uppercase\expandafter{\romannumeral4}.B.

\begin{figure*}[!t]
\includegraphics[width=\linewidth,height=70pt]{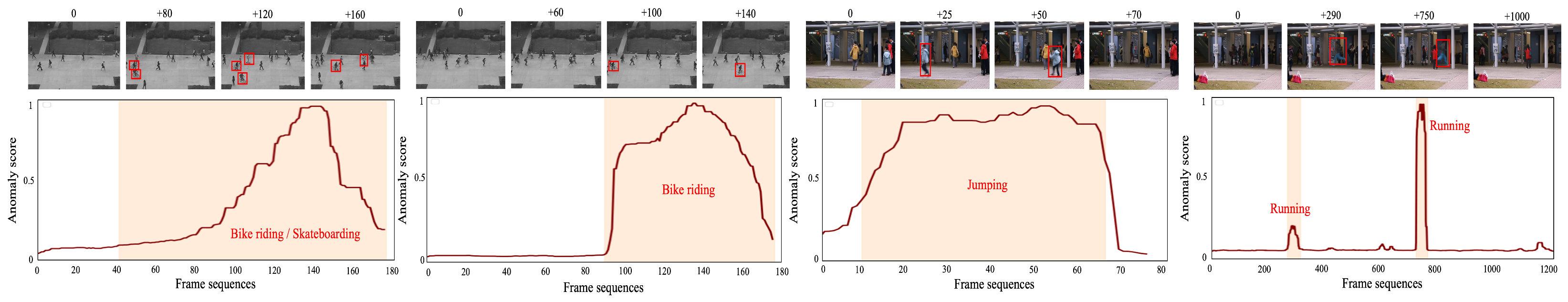}
\centering
\caption{
Illustrations of anomaly score that denotes the reconstruction error in Ped2(first, second column), and Avenue(third, fourth column), datasets. Red bounding boxes in frames represent the abnormal objects. Orange region in graph denotes the time sequences that abnormal situation exists in video frames. As shown in graph, anomaly scores dramatically increase with the high reconstruction error when the abnormal frames start.}
\label{fig_sim}
\end{figure*}

\section{Experimental Evaluations} 
\subsection{Datasets}
UCSD Ped2\cite{mahadevan2010anomaly} consists of 16 training videos and 12 testing videos with the same background. It contains content that shows people riding bicycles and skateboards and driving trucks as abnormal events. 

CUHK Avenue\cite{lu2013abnormal} consists of 16 training and 21 testing videos with the same background. It has more complex and a larger number of abnormal behaviors than Ped2, e.g., a running person, littering, throwing a bag or paper, jumping, etc. 

ShanghaiTech\cite{luo2017revisit} consists of 330 training and 107 testing videos with different backgrounds. It has the largest amount of datasets and challenging datasets with various abnormality situations.

\subsection{Implementation details}
Inspired by \cite{liu2021hybrid}, the proposed approach extracts foreground objects from each frame, and constructs a cube that contains the current frame and the previous four frames with the same coordinates. The size of the object-level frame is $3\times32\times32$ pixels. Similarly, object-level optical flows are extracted using FlowNet2.0\cite{ilg2017flownet}.
We adopt the AdamW optimizer with
\begin{math}lr=1.5e-4\end{math}, \begin{math}wd=0.05\end{math}, \begin{math}eps=1e-8\end{math}, and \begin{math}\beta_{1}=0.9,\beta_{2}=0.95\end{math}. Also, we adopt the CosineAnnealingWarmRestarts\cite{loshchilov2016sgdr} as learning scheduler with \begin{math}min\_lr=1e-5\end{math}.
Previously, we define the weight of the anomaly score using \begin{math}\lambda_{a}\end{math} and \begin{math}\lambda_{o}\end{math}.
During the test, we apply \begin{math}(\lambda_{a},\lambda_{o})=(0.05,0.94)\end{math} in Ped2, \begin{math}(\lambda_{a},\lambda_{o})=(2.0,1.0)\end{math} in Avenue and SHTech.
Similar to the previous work \cite{liu2021hybrid}, we empirically selected the parameters for the experiment. 
The batch size of all datasets is 128, and the masking ratio of Ped2 and Avenue/SHTech is 0.5 and 0.75, respectively.

\subsection{Quantitative and qualitative evaluation}
We compare our model with \cite{gong2019memorizing,liu2021hybrid,he2021masked} using object-level appearance frames on Avenue dataset.
MemAE\cite{gong2019memorizing} places one memory module between the encoder and the decoder.
In contrast, ML-MemAE-SC in \cite{liu2021hybrid}, includes multi-level memory modules with skip connections.
We experiment with these modules using Avenue dataset, following reconstruction-based methods.

\begin{table}[!t]
\renewcommand{\arraystretch}{1.3}
\setlength{\tabcolsep}{10pt}
\caption{AUROC performance (\%) comparison for object-level VAD task using appearance frames}
\label{table_example}
\centering
\begin{tabular}{c|cccc}
\Xhline{2\arrayrulewidth}
Method & MemAE & ML-MemAE-SC & MAE & Ours\\
\Xhline{2\arrayrulewidth}
AUROC & 70.2 & 81.9 & 81.6 & \textbf{86.2} \\
\Xhline{2\arrayrulewidth}
\end{tabular}
\end{table}

\begin{table}[!t]
\renewcommand{\arraystretch}{1.3}
\setlength{\tabcolsep}{14pt}
\caption{AUROC performance (\%) comparison for motion module. Our framework shows the best performance when adopting the CAE module.}
\label{table_example}
\centering
\begin{tabular}{c|ccc}
\Xhline{2\arrayrulewidth}
Method & MemAE & ML-MemAE-SC & CAE\\
\hline
Flow & 77.9 & 86.8 & 79.8\\
Overall & 89.3 & 91.7 & \textbf{92.1}\\
\Xhline{2\arrayrulewidth}
\end{tabular}
\end{table}

\newcommand{\cmark}{\ding{51}}%
\newcommand{\xmark}{\ding{55}}%
\begin{table}[!t]
\renewcommand{\arraystretch}{1.3}
\setlength{\tabcolsep}{10pt}
\caption{AUROC performance (\%) for ablation study}
\label{table_example}
\centering

\begin{tabular}{ccc|c}
\Xhline{2\arrayrulewidth}
masked & whole & partial & result \\
\hline
 \xmark & \xmark & \xmark &  \textbf{70.5} \\
 
 \cmark & \xmark & \xmark &  \textbf{81.6} \\

 \cmark & \cmark & \xmark &  \textbf{84.9} \\

 \cmark & \cmark & \cmark &  \textbf{86.2} \\

\Xhline{2\arrayrulewidth}
\end{tabular}
\end{table}

In addition, we use 
memory items with 2000 and 100 slots in each \cite{gong2019memorizing,liu2021hybrid}, which show better performance than the other numbers of slots.
As shown in Fig. 4, we visualize the reconstruction error maps of some abnormal and normal events. We show that in our model, the difference between the reconstruction error of the abnormal events and that of the normal events is clearly larger than recent AE with memory modules, ML-MemAE-SC in \cite{liu2021hybrid}. 
Thus, we conclude that our contextual-appearance module distinguishes abnormal and normal events well, alleviating the \textit{generalizing issue} since this issue causes a small difference between the reconstruction error of the normal data and that of the abnormal data.

As shown in TABLE \uppercase\expandafter{\romannumeral1}, MemAE exhibits a low performance with 70.2\%, whereas the improved MemAE, ML-MemAE-SC exhibits better performance with 81.9\%. However, both are lower than the performance of our model, which achieves a result of 86.2\%. 

\subsection{Ablation study}
To verify the effectiveness of our contextual-appearance module, a comprehensive analysis is conducted on the Avenue dataset. We examine our model on the four different settings: 1) none, 2) masked, 3) masked \& whole, 4) all.
\subsubsection{None}
We firstly assess the model with none of the proposed prediction methods applied. It is identical to the existing future frame prediction methods of predicting a single future frame. We observe that the VAD performance(AU-ROC) in this vanilla setting is only 70.5\%. It can be interpreted as that predicting a single future frame is not enough for model to learn about the context of the video, as the video is composed of the changes of continuous frames.

\begin{figure}[!t]
\centering
\includegraphics[width=3.3in]{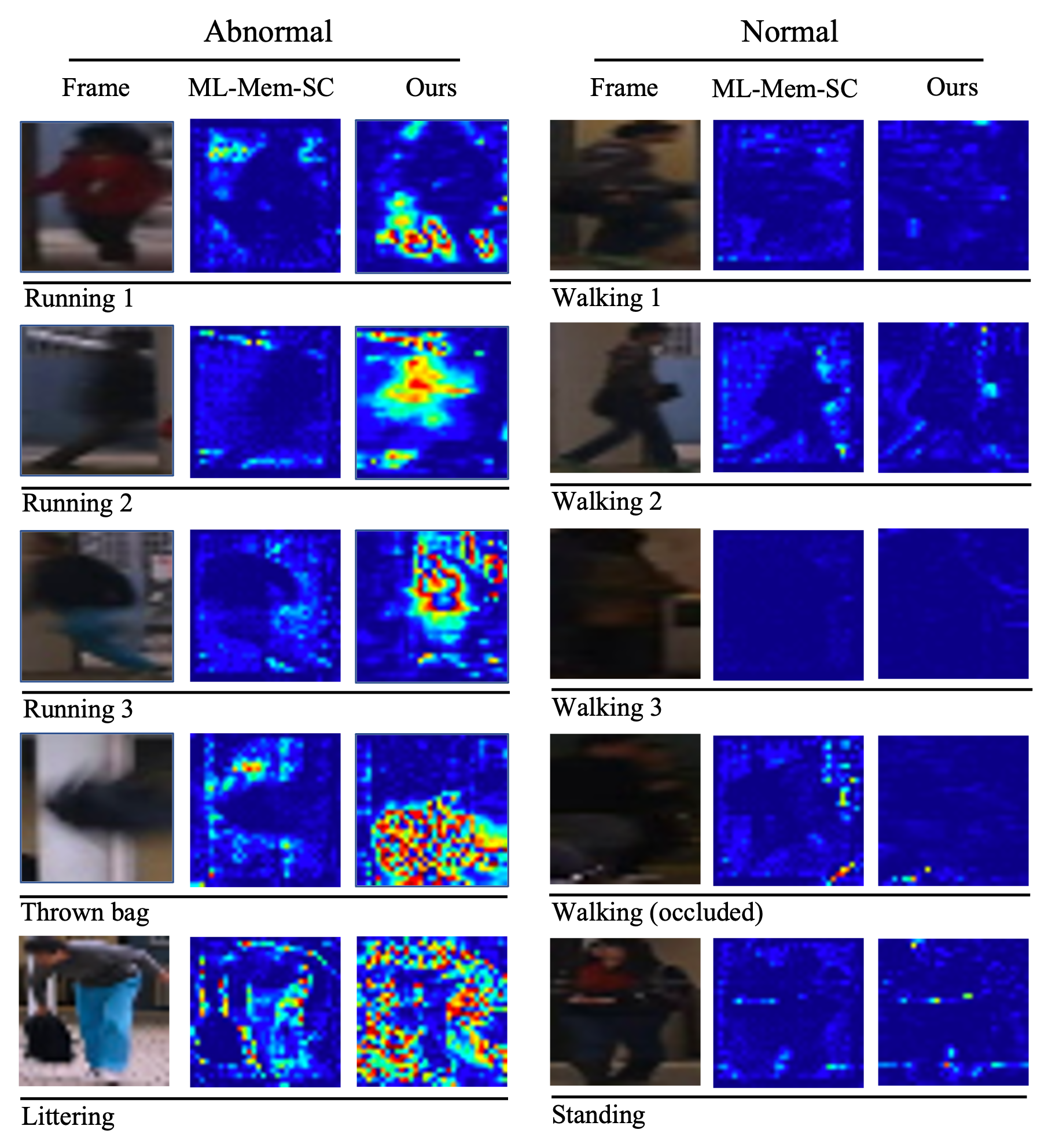}
\caption{Reconstruction error maps of the abnormal and normal behaviors are shown in the left and right region of the figure, respectively. The left columns of each region show appearance frames, the middle columns show reconstruction error maps of ML-MemAE-SC in \cite{liu2021hybrid}, and the right columns show reconstruction error maps of our contextual-appearance module. }
\label{fig_sim}
\end{figure}

\subsubsection{Masked}
TABLE \uppercase\expandafter{\romannumeral3} shows that simultaneously predicting the missing frames among the continuous frames improves the VAD performance with a great margin of 11.1\%. This is due to the improvement of understanding of spatio-temporal context of the multiple frames which is achieved by predicting the masked frames. Moreover, as our model calculates the final error of the simultaneous prediction by adding up all the reconstruction error of each frame, the discrepancy between the normal and the abnormal frames becomes more distinct than the single frame prediction based methods. We argue that this evaluation process contributes to the alleviation of the  \textit{generalizing issue}.

\subsubsection{Masked \& whole}
Integrating the whole future frame prediction improves 3.3\% of VAD performance compared to the situation of predicting the masked frames only. We attribute this performance improvement to two main reasons:
(1) Compared to the reconstruction-based methods which generalize the input frame as it is given, future frame prediction is less sensitive to the \textit{generalizing issue} as the prediction to the unseen frame is made based only on the previous frames. 
(2) In the whole future prediction, the predicted frame can effectively reflect the normality as the prediction is conducted based on the masked tokens composed of the parameters that are trained only on the normal data and the non-masked tokens.

\subsubsection{All}
Lastly, we apply the partial future prediction to the situation of predicting both of the masked and whole frames, and observe 1.3\% VAD performance improvement. It can be interpreted as that the latent feature from the encoder gets more elaborated, while performing a more challenging task of predicting the next frame based only upon the restricted information during the training.
\subsection{Motion module}
To examine the complementary effect with our contextual-appearance module and the motion information, we apply optical flow samples to CAE and other memory module based approaches(MemAE\cite{gong2019memorizing}, ML-MemAE-SC in \cite{liu2021hybrid}). 
The memory used in the \cite{gong2019memorizing}, \cite{liu2021hybrid} consists of 2000 slots. 
when combined with our contextual-appearance module in TABLE \uppercase\expandafter{\romannumeral2}, CAE yields the best VAD performance of 92.1\%. Based on this finding, the final conclusion can be drawn: the model can detect anomalies more effectively when the flow reconstruction error is added to the anomaly score, compared to when only either of the flow or contextual-appearance module is used. It implies that the anomaly scores of the flow and appearance complement each other well.

 \subsection{Results}
As shown in TABLE \uppercase\expandafter{\romannumeral4}, we compare our model with other previous VAD approaches which can be classified into three main categories: reconstruction-based, prediction-based, and the hybrid methods. First, reconstruction-based methods include Conv-AE\cite{hasan2016learning}, 3D-Conv\cite{zhao2017spatio}, MemAE\cite{gong2019memorizing}, and MNAD-R\cite{park2020learning}, while MNAD-P\cite{park2020learning}, AMMC-Net\cite{cai2021appearance}, Frame-Pred\cite{liu2018future}, VEC\cite{yu2020cloze}, C2-D2GAN\cite{feng2021convolutional}, and Transanomaly\cite{yuan2021transanomaly}, are prediction-based methods and  HF2-VAD\cite{liu2021hybrid} is the hybrid method.
We observe that our model achieves better results than other state-of-the-art methods in Ped2, except for HF2-VAD\cite{liu2021hybrid}.
HF2-VAD\cite{liu2021hybrid} captures the high correlation between appearance and optical flow to predict the next frame, thus the result is highly dependent on the optical flow.
On the other hand, as our model trains each module separately, and then integrates the score from each module, the optical flow score could be balanced out by the appearance score. It means that the offset might happen for the effects to disappear as they affect each other.
Hence, HF2-VAD\cite{liu2021hybrid} show superior performance on Ped2, as most of the Ped2 define abnormal events simply based on the speed.
On the other hand, our model surpasses all the other baselines on Avenue dataset and shows competitive performance on SHtech dataset as well.
We use the area under the receiver operating characteristic curve (AUROC) as the evaluation metric.
Furthermore, as shown in Fig. 3, we also visualize anomaly score graphs of Ped2, Avenue. We show that the scores of the abnormal behaviors are larger than the normal counterpart. 
 
\begin{table}[!t]
\renewcommand{\arraystretch}{1.3}
\caption{AUROC performance (\%) comparison with previous works on UCSD Ped2, CUHK Avenue and SHTech datasets}
\label{table_example}
\centering
\begin{tabular}{c|ccc}
\Xhline{2\arrayrulewidth}
Method & UCSD Ped2 & CUHK Avenue & SHTech\\
\hline
Conv-AE \cite{hasan2016learning} & 85.0 & 80.0 & -\\ 

3D-Conv \cite{zhao2017spatio} & 91.2 & 80.9 & - \\

MemAE \cite{gong2019memorizing} & 92.1 & 83.3 & 71.2\\

MNAD-R \cite{park2020learning} & 90.2 & 82.8 & 69.8 \\
\hline
MNAD-P \cite{park2020learning} & 97.0 & 88.5 & 70.5\\

AMMC-Net \cite{cai2021appearance} & 96.6 & 86.6 & 73.7\\

Frame-Pred \cite{liu2018future} & 95.4 & 85.1 & 72.8\\

VEC \cite{yu2020cloze} & 97.3 & 90.2 & 74.8\\

CT-D2GAN \cite{feng2021convolutional} & 97.2 & 80.9 & 77.7\\

TransAnomaly \cite{yuan2021transanomaly} & 96.4 & 87.0 & -\\
\hline
HF2-VAD \cite{liu2021hybrid} & 99.3 & 91.1 & 76.2 \\
\hline
Ours &  \textbf{98.0} &  \textbf{92.1} &  \textbf{75.3}\\
\Xhline{2\arrayrulewidth}
\end{tabular}
\end{table}

\section{Conclusion}
In this study, we achieve comparable results to state-of-the-art VAD performance by proposing a ViT-based framework that learns the multi-contextual feature representations. Our training mechanism simultaneously predicts the multiple components of consecutive normal frames, resulting in alleviating \textit{generalizing issue} for VAD.
In addition, we successfully apply the ViT model to the anomaly detection domain by designing a single frame to be processed as one patch, which leads to perform well on the VAD task with maintaining the advantages of the transformer architecture. 
In verified experimental settings, we demonstrate that our contextual-appearance module creates synergy with the CAE motion module, resulting in a competitive performance compared to the existing methods, including a large discrepancy between reconstruction errors of normal and abnormal events.

\section*{Acknowledgment} 
We thank Sueyeon Kim for proofreading the manuscript and related discussions. This work was conducted by Center for Applied Research in Artificial Intelligence(CARAI) grant funded by Defense Acquisition Program Administration(DAPA) and Agency for Defense Development(ADD) (UD190031RD).






%

\bibliographystyle{./IEEEtran}
\bibliography{./IEEEabrv,./gu}

\end{document}